\documentclass[lettersize,journal]{IEEEtran}
\usepackage{amsmath,amsfonts}
\usepackage{algorithmic}
\usepackage{algorithm}
\usepackage{array}
\usepackage[caption=false,font=normalsize,labelfont=sf,textfont=sf]{subfig}
\usepackage{textcomp}
\usepackage{stfloats}
\usepackage{url}
\usepackage{verbatim}
\usepackage{graphicx}
\usepackage{cite}
\usepackage{amsmath,amsfonts,bm}

\def\eqref#1{equation~\ref{#1}}
\def\1{\bm{1}}

\DeclareMathAlphabet{\mathsfit}{\encodingdefault}{\sfdefault}{m}{sl}
\SetMathAlphabet{\mathsfit}{bold}{\encodingdefault}{\sfdefault}{bx}{n}

\usepackage{multirow}
\usepackage{booktabs}
\usepackage{wrapfig}
\usepackage{amssymb}
\usepackage{diagbox}
\usepackage{bbding}
\usepackage{xr}
\usepackage{tikz}
\usepackage[hidelinks]{hyperref}
\usepackage{orcidlink}

\newcommand{\firstone}[1]{\colorbox{red!15}{#1}}
\newcommand{\secondone}[1]{\colorbox{blue!15}{#1}}
\definecolor{learnable_color}{RGB}{251,229,214}

\newcommand{\email}[1]{\href{mailto:#1}{#1}}

\begin{document}

\title{Temporal-Guided Visual Foundation Models for Event-Based Vision}


\author{
    Ruihao Xia~\orcidlink{0009-0008-6749-5104}, 
    Junhong Cai~\orcidlink{0009-0002-7527-5028},
    Luziwei Leng~\orcidlink{0000-0002-9344-8589}, \emph{Senior Member, IEEE},
    Liuyi Wang~\orcidlink{0000-0003-1368-0300}, 
    Chengju Liu~\orcidlink{0000-0001-7543-0855}, \\ \emph{Member, IEEE}, 
    Ran Cheng~\orcidlink{0000-0001-9410-8263}, \emph{Senior Member, IEEE}, 
    Yang Tang~\orcidlink{0000-0002-2750-8029}, \emph{Fellow, IEEE},
    and Pan Zhou~\orcidlink{0000-0003-3400-8943}

\thanks{This work was supported by National Natural Science Foundation of China (Grants No. 62233005, U2441245), and Interdisciplinary Innovation and Education Integration Project under the Shanghai Municipal Peak Discipline Program in Intelligent Science and Technology (Category IV). \emph{(Corresponding authors: Luziwei Leng; Yang Tang.)}}
\thanks{Ruihao Xia, and Yang Tang are with the Key Laboratory of Smart Manufacturing in Energy Chemical Process, Ministry of Education, East China University of Science and Technology, Shanghai 200237, China  (e-mail: \email{xia\_rho@mail.ecust.edu.cn}; \email{yangtang@ecust.edu.cn}).}
\thanks{Luziwei Leng is with ACSLab, Huawei Technologies Company Ltd., Shenzhen 518055, China. (e-mail: \email{lengluziwei@huawei.com}).}
\thanks{Junhong Cai is with the Department of Computer Science and Engineering, Southern University of Science and Technology, Shenzhen 518055, China. (e-mail: \email{12332479@mail.sustech.edu.cn}).}
\thanks{Liuyi Wang, and Chengju Liu are with Shanghai Institute of Intelligent Science and Technology, Tongji University, Shanghai 201210, China, and also with State Key Laboratory of Autonomous Intelligent Unmanned Systems, Tongji University, Shanghai 201210, China. (e-mail: \email{wly@tongji.edu.cn}; \email{liuchengju@tongji.edu.cn}).}
\thanks{Ran Cheng is with the Department of Data Science and Artificial Intelligence and the Department of Computing, The Hong Kong Polytechnic University, Hong Kong SAR 999077, China. (e-mail: \email{ranchengcn@gmail.com}).}
\thanks{Pan Zhou are with Singapore Management University, 188065, Singapore (e-mail: \email{panzhou@smu.edu.sg}).} 
% \thanks{This work was done during Ruihao Xia’s internship at Huawei.} 
}



\maketitle

\begin{abstract}
Event cameras offer unique advantages for vision tasks in challenging environments, yet processing asynchronous event streams remains an open challenge. While existing methods rely on specialized architectures or resource-intensive training, the potential of leveraging modern Visual Foundation Models (VFMs) pretrained on image data remains under-explored for event-based vision. To address this, we propose Temporal-Guided VFM (TGVFM), a novel framework that integrates VFMs with our temporal context fusion block seamlessly to bridge this gap. Our temporal block introduces three key components: (1) Long-Range Temporal Attention to model global temporal dependencies, (2) Dual Spatiotemporal Attention for multi-scale frame correlation, and (3) Deep Feature Guidance Mechanism to fuse semantic-temporal features. By retraining event-to-video models on real-world data and leveraging transformer-based VFMs, TGVFM preserves spatiotemporal dynamics while harnessing pretrained representations. Experiments demonstrate SoTA performance across semantic segmentation, depth estimation, and object detection, with improvements of 16\%, 21\%, and 16\% over existing methods, respectively. Overall, this work unlocks the cross-modality potential of image-based VFMs for event-based vision with temporal reasoning. Code is available at \url{https://github.com/XiaRho/TGVFM}.
\end{abstract}


\section{Introduction}
\IEEEPARstart{E}{vent} cameras~\cite{EventCameras1,TCSVT_Classification} capture per-pixel brightness changes asynchronously, offering distinct advantages over conventional frame-based imaging, including high dynamic range, microsecond-level temporal resolution, and low power consumption~\cite{TCSVT_Learning,EventCamerasSurvey}. These unique properties make event cameras highly suitable for vision tasks in challenging environments, such as high-speed motion~\cite{TCSVT_Deblurr,TCSVT_Track,EventHighSpeed} and extreme lighting conditions~\cite{TCSVT_Light,EventLowLight}.  
 
Despite these advantages, the effective processing of event streams remains an open challenge. Recent research has primarily focused on designing specialized network architectures~\cite{HMNet,EReFormer} or developing resource-intensive training pipelines~\cite{TCSVT_Distill,ECDP,ECDDP} to handle event-based data. While these methods have achieved promising results, they require extensive engineering efforts and large-scale annotated event datasets, limiting their scalability and adaptability.  In contrast, Visual Foundation Models (VFMs), e.g., Rein~\cite{Rein} for semantic segmentation and Metric3D~\cite{Metric3D} for monocular depth estimation, have shown remarkable generalization capabilities across diverse vision tasks. These models, pretrained on massive image datasets, exhibit strong transferability and adaptability to unseen scenarios, often surpassing traditional task-specific networks in both efficiency and performance. However, VFMs remain largely unexplored in the event-based domain. Successfully adapting VFMs for event-based vision could significantly advance the field by leveraging their powerful pretrained representations while reducing the reliance on complex architectures and labor-intensive training.  

\begin{figure}[t] 
    % \vspace{0pt}
    \centering
    \includegraphics[width=0.49\textwidth]{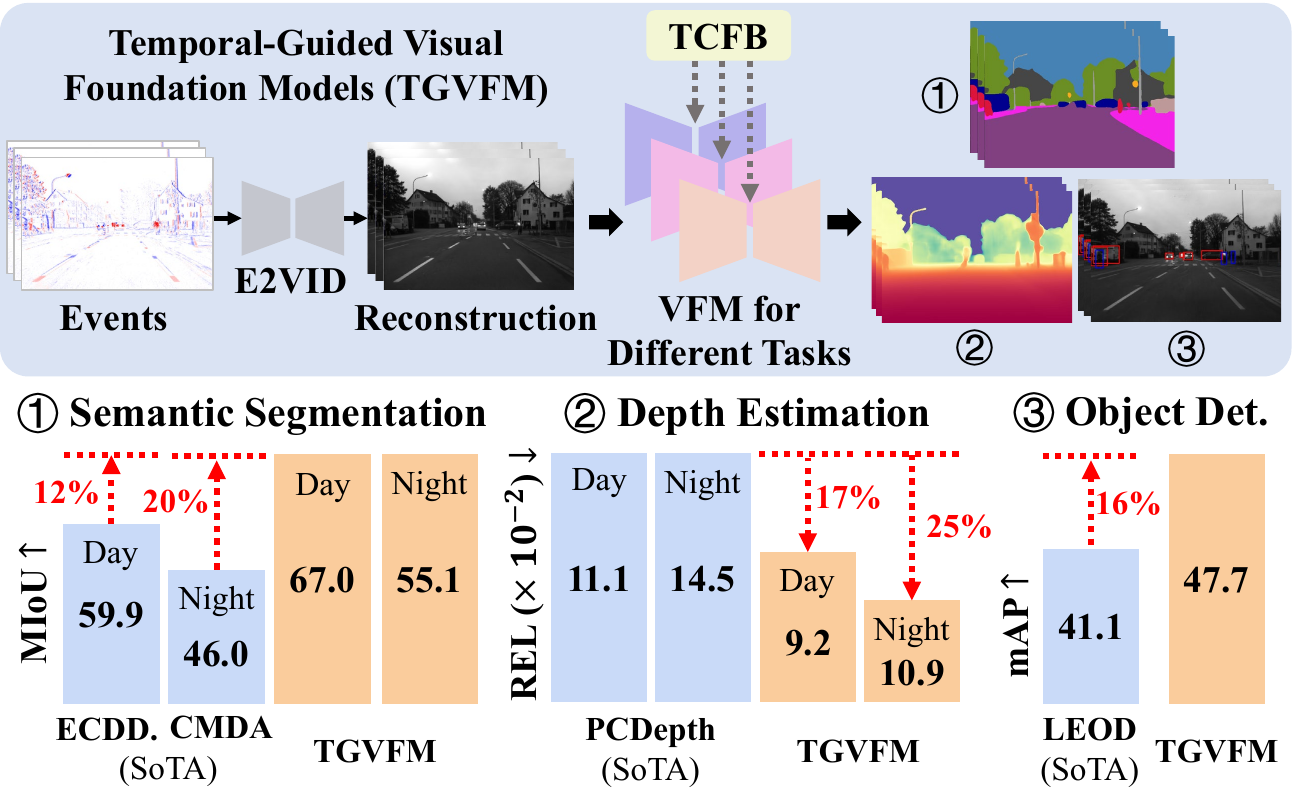}
    % \vspace{-8pt}
    \caption{
        (1) TGVFM: Our proposed Temporal Context Fusion Block (TCFB) is integrated in a unified manner into VFMs specifically designed for different tasks, extending the spatial reasoning capability of traditional VFMs to spatio-temporal reasoning. (2) Experiments: Compared to the SoTA methods ECDDP~\cite{ECDDP}, CMDA~\cite{CMDA}, PCDepth~\cite{PCDepth}, and LEOD~\cite{LEOD} in the day and night sequences of DSEC datasets~\cite{DSEC,ESS,CMDA,DSEC-Det}, our TGVFM demonstrates significant improvements in all tasks.
    }
    \label{fig:motivation}
    % \vspace{-12pt}
\end{figure}

An early attempt to bridge VFMs and event-based vision is Events-to-Video (E2VID)~\cite{E2VID,E2VID2,TCSVT_E2VID}, which reconstructs synthetic grayscale frames from event streams using a recurrent network and then processes these frames with conventional CNN-based VFMs. 
However, the direct integration of E2VID and VFMs suffers from a fundamental limitation. 
VFMs are designed for static image data, in which spatial structures dominate feature extraction. When applied to reconstructed synthetic frames, these models lose access to the intricate temporal dependencies of event streams, leading to suboptimal performance. 
This limitation highlights the need for a more effective approach that preserves temporal information while harnessing the generalization power of VFMs.  

% \noindent\textbf{Contribution.} 

In this work, we revisit the integration of VFMs into event-based vision to address the above limitations. Specifically, we propose a novel \textit{Temporal Context Fusion Block (TCFB)} designed for efficient integration into transformer-based VFMs and build our Temporal-Guided VFM (TGVFM) framework, as shown in Figure~\ref{fig:motivation}.  
Our TCFB hierarchically models spatiotemporal dependencies within continuous event-derived frames. This plug-and-play architecture substantially preserves VFM pretrained knowledge while enabling dynamic temporal reasoning through three core components. 
(1) \textbf{Long-Range Temporal Attention (LTA)}: This module introduces a memory-augmented self-attention mechanism, where each token in the current frame interacts with its historical counterparts stored in a sliding memory bank through temporal self-attention.  This design explicitly maintains positional correspondence across time steps while aggregating global temporal context, allowing the model to retain long-term dependencies.  
(2) \textbf{Dual Spatiotemporal Attention (DSA)}: To bridge adjacent frames at varying scales, we deploy dual attention pathways. Specifically, inter-frame cross-attention projects queries from previous frame features against keys/values of the current frame, establishing direct temporal correspondence between feature anchors. Local window self-attention operates within spatiotemporal windows that span consecutive frames, enabling fine-grained interaction between a token and its temporal neighbors.
(3) \textbf{Deep Feature Guidance Mechanism (DFGM)}: Recognizing the predictive significance of high-level semantic features, we reuse previous embeddings from deeper network layers in the context of temporal and local window self-attention. Specifically, these semantically rich features are adaptively fused with shallow features of previous frames via learnable patch embeddings. Then, self-attention operations are carried out on fused semantic features from previous frames and non-semantic current-frame features. 

Furthermore, we systematically analyze the quality of synthetic frames generated by E2VID and find that prior E2VID models~\cite{E2VID}, trained on low-quality synthetic datasets from event simulators~\cite{ESIM}, produce degraded grayscale reconstructions~\cite{ESS}. To address this, we retrain E2VID on high-fidelity real-world datasets (DSEC~\cite{DSEC}), significantly improving the perceptual quality of event-derived frames.
Additionally, previous methods primarily relied on CNN-based VFMs~\cite{OCRNet,Deeplab}, which exhibit limited cross-modality adaptability compared to transformer-based VFMs~\cite{Rein,Metric3D,Swin}. To leverage their strengths, we integrate our proposed temporal context fusion block with more robust transformer-based VFMs, developing TGVFM, which enhances generalization while incorporating temporal perception.   

Finally, extensive experimental results show that our framework achieves state-of-the-art (SoTA) performance across several event-based tasks as illustrated in Figure~\ref{fig:motivation}.
On semantic segmentation, depth estimation, and object detection, our method achieves improvement over the corresponding SoTAs by a significant 16\%, 21\%, and 16\%, respectively. These results demonstrate the feasibility of our TGVFM by integrating VFMs with the proposed temporal context fusion block for event-based vision, offering a scalable and efficient alternative to conventional event processing pipelines.  

\noindent\textbf{Contributions.} Our main contributions are summarized:
\begin{itemize}
    \item We revisit the integration of VFMs into event-based vision and identify the temporal limitations of prior E2VID-based pipelines.  
    \item We propose a \textit{Temporal Context Fusion Block (TCFB)} that enables transformer-based VFMs to model long-range temporal dependencies while preserving pretrained spatial knowledge.  
    \item Our Temporal-Guided VFM (TGVFM) achieves SoTA performance on multiple event-based benchmarks.  
\end{itemize}

\section{Related Works}
\label{sec:related}

\subsection{Event-based Vision}
Recent advances in event-based vision have pursued performance gains through three primary avenues: specialized pretraining, complex network architectures, and meticulously designed optimization.

\noindent \textbf{Self-Supervised Pretraining.} Methods like ECDP~\cite{ECDP} and ECDDP~\cite{ECDDP} address the scarcity of labeled event data through contrastive learning frameworks. ECDP introduces event-specific data augmentations and cross-modal alignment between synthetic event frames and RGB images, employing an embedding projection loss to prevent model collapse. ECDDP tackles event data sparsity by clustering event patch features and enforcing context-to-context similarity relationships.

\noindent \textbf{Architectural Innovations.} HMNet~\cite{HMNet} pioneers a hierarchical memory architecture with multi-rate latent states to encode dynamic scene contents across temporal scales. PCDepth~\cite{PCDepth} discretizes scenes into high-level patterns for complementary learning between event data and images. EReFormer~\cite{EReFormer} integrates recurrent mechanisms into vision transformers, leveraging GRViT modules to model long-term temporal dependencies in event streams.

\noindent \textbf{Optimization-Centric Designs.} Building on architectural innovations, OpenESS~\cite{OpenESS} bridges image-text CLIP knowledge to event streams via frame-to-event contrastive distillation and semantic consistency regularization, enabling annotation-efficient segmentation. LEOD~\cite{LEOD} introduces a self-training paradigm with bi-directional inference and tracking-based pseudo-label refinement for semi-supervised object detection.

These approaches collectively highlight the field's emphasis on modality-specific customization. However, they overlook opportunities to leverage pretrained knowledge from image-based VFMs. Our work addresses this gap through temporal-aware adaptation of pretrained robust VFMs.

\subsection{Visual Foundation Models}
The evolution of VFMs has significantly influenced modern computer vision. Early CNN-based architectures like ResNet~\cite{ResNet}, trained on ImageNet-1K~\cite{ImageNet}, established the paradigm of transfer learning through supervised pretraining. However, their generalization to unseen scenarios remained constrained by limited model capacity and dataset diversity~\cite{OCRNet,Deeplab}. The advent of vision transformers~\cite{ViT} marked a pivotal shift, with architectures like Swin Transformer~\cite{Swin} introducing hierarchical attention mechanisms and improved inductive biases for dense prediction tasks. These transformer-based models demonstrated enhanced robustness across domains by capturing long-range dependencies and multi-scale features.

Recent breakthroughs in self-supervised learning on large-scale data further unlocked unprecedented generalization. DINOv2~\cite{DINOv2}, pretrained on the LVD-142M dataset via self-supervision, learns universal visual representations that transfer effectively to diverse downstream tasks. Such advancements highlight the potential of large-scale pretraining in developing dataset-agnostic VFMs. 

Despite these strides, existing VFM research predominantly focuses on conventional image modality, overlooking event-based vision. Early attempts to bridge this gap relied on CNN-based VFMs applied to reconstructed event frames~\cite{E2VID}, but suffered from suboptimal reconstruction quality and limited model adaptability~\cite{ESS}. The emergence of robust transformer-based VFMs, coupled with retrained high-fidelity event-to-frame conversion, creates new avenues for re-purposing image modality knowledge in event-based vision. Our work capitalizes on this new direction through a systematic exploration.

\begin{figure*}[t]
    \centering
    \includegraphics[width=0.99\textwidth]{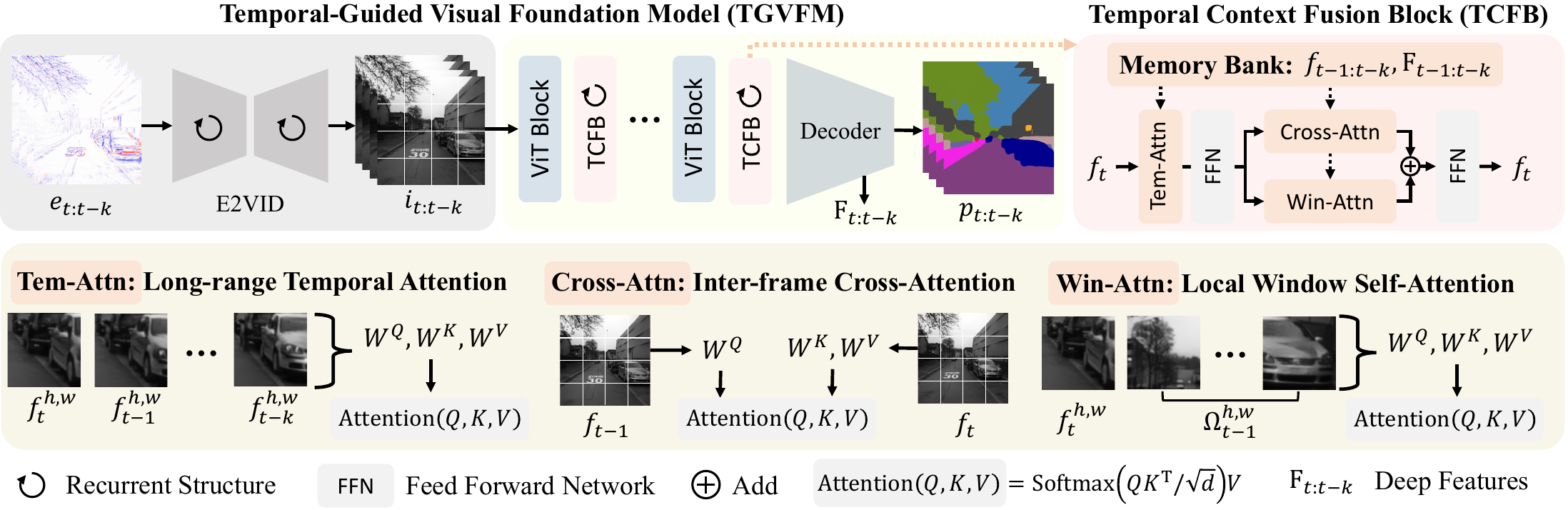}
    % \vspace{-8pt}
    \caption{
        Our TGVFM framework integrates several proposed TCFB between ViT blocks to extract both spatial and temporal features among multiple frames. In each TCFB, the input feature $f_t$ processed by different attention operations to interact with previous features $f_{t-1:t-k}$ and $\mathbf{F}_{t-1:t-k}$ stored in the memory bank for temporal reasoning. For clarity, we omit the residual connections in the attention and feed-forward network.
    }
    \label{fig:framework}
    % \vspace{-5pt}
\end{figure*}

\section{Methodology}
\label{sec:method}

\subsection{Motivation and Overall Framework}
As a pioneer, E2VID~\cite{E2VID,E2VID2} bridges VFMs and event-based vision via a two-stage approach: grayscale reconstruction via E2VID followed by VFM processing.  Given events within a temporal window (e.g., 50ms), they are converted into a voxel grid~\cite{VoxelGrid} \( e_t \! \in\! \mathbb{R}^{H\times W\times C} \), where \( C \) is the temporal channel. Then, E2VID \( f_{\mathrm{E2VID}} \) reconstructs grayscale with recurrent state propagation:  
\begin{equation}
i_t, s_t = f_{\mathrm{E2VID}}(e_t, s_{t-1}),
\end{equation}
where $s_t$ is the temporal state in recurrent modules like ConvLSTM~\cite{ConvLSTM}, and \( i_t \in \mathbb{R}^{H\times W} \) is the reconstructed grayscale frame. Subsequently, \( i_t \) is fed into the VFM \( f_{\mathrm{VFM}} \) for predictions:  
\begin{equation}
p_t = f_{\mathrm{VFM}}(i_t).
\end{equation}

However, the above process suffers from a critical limitation: conventional VFMs are predominantly designed for single-frame inputs, processing consecutive frames independently. This paradigm fails to explicitly model temporal dependencies between adjacent reconstructed frames, which are pivotal for event-based vision.

To address this limitation, we propose the Temporal-Guided Visual Foundation Model (TGVFM), as shown in Figure~\ref{fig:framework}. Our framework operates in two phases. First, following E2VID, we convert event streams into continuous grayscale frames. However, instead of relying on the original model, we retrain E2VID on real-world datasets for high-fidelity reconstruction, as detailed in Section~\ref{otherimprovement}.  

Second, we introduce the Temporal Context Fusion Block (TCFB), a novel module seamlessly integrated into transformer-based VFMs to enable continuous spatiotemporal reasoning. The enhanced TGVFM \( f_{\mathrm{TGVFM}} \) leverages a memory bank \( \mathcal{M} \) to retain multi-scale spatiotemporal features across network stages. Our temporal block enables dynamic feature propagation and cross-stage fusion, leading to improved temporal modeling while maintaining the strong generalization of pretrained VFMs—without requiring specialized architectures or extensive retraining:  $p_t = f_{\mathrm{TGVFM}}(i_t, \mathcal{M}).$ Figure~\ref{fig:framework} illustrates the three synergistic components of our TCFB:  
1) Long-Range Temporal Attention (LTA), 2) Dual Spatiotemporal Attention (DSA), and  3) Deep Feature Guidance Mechanism (DFGM). Each of these components is introduced in the following sections.

\subsection{Long-Range Temporal Attention (LTA)}

Traditional VFMs process each frame independently, inherently disregarding the temporal continuity of event-derived frames. While naive temporal concatenation or averaging can aggregate multi-frame features, such methods fail to model long-range dependencies or preserve positional correspondence—critical for capturing coherent motion trajectories and transient patterns in event data. To address this, we design LTA to enable position-aware global temporal reasoning: each spatial token dynamically attends to its historical counterparts across an extended time horizon. This mechanism allows pretrained VFMs to retain their spatial attention priors while learning to correlate temporally distant but semantically consistent regions, effectively bridging the gap between static image understanding and event-based temporal dynamics.

Specifically, the LTA module establishes global temporal correspondence through temporal-dimensional self-attention. For each spatial location $(h,w)$ in feature map $f_t\in \mathbb{R}^{H\times W\times C}$, we construct query, key, and value vectors by linearly projecting its temporal sequence:
\begin{equation}
\label{equ:temporal_qkv}
Q = W^{Q} \cdot f^{h,w}_{t:t-k},~
K = W^{K} \cdot f^{h,w}_{t:t-k},~
V = W^{V} \cdot f^{h,w}_{t:t-k},
\end{equation}
where $f^{h,w}_{t:t-k}=[f^{h,w}_{t},\ldots,f^{h,w}_{t-k}] \in \mathbb{R}^{k\times C}$ contains historical features at the same spatial coordinate, and $W^{Q},W^{K},W^{V} \in \mathbb{R}^{C\times d}$ are learnable projection matrices. The temporal attention updates features through:
\begin{equation}
\label{equ:attenion}
\hat{f}^{h,w}_t = \mathrm{Softmax}\left(\frac{QK^\top}{\sqrt{d}}\right)V + f^{h,w}_t,
\end{equation}
where $d$ is the output dimension of query and key features.
This formulation achieves critical objectives through adaptive temporal weighting by attention scores that reflect inter-frame relevance. The sliding memory window $k$ balances computational efficiency with long-range context capture.

\begin{figure*}[t]
    \centering
    \begin{minipage}{0.56\textwidth} 
        \centering
        \includegraphics[width=\linewidth]{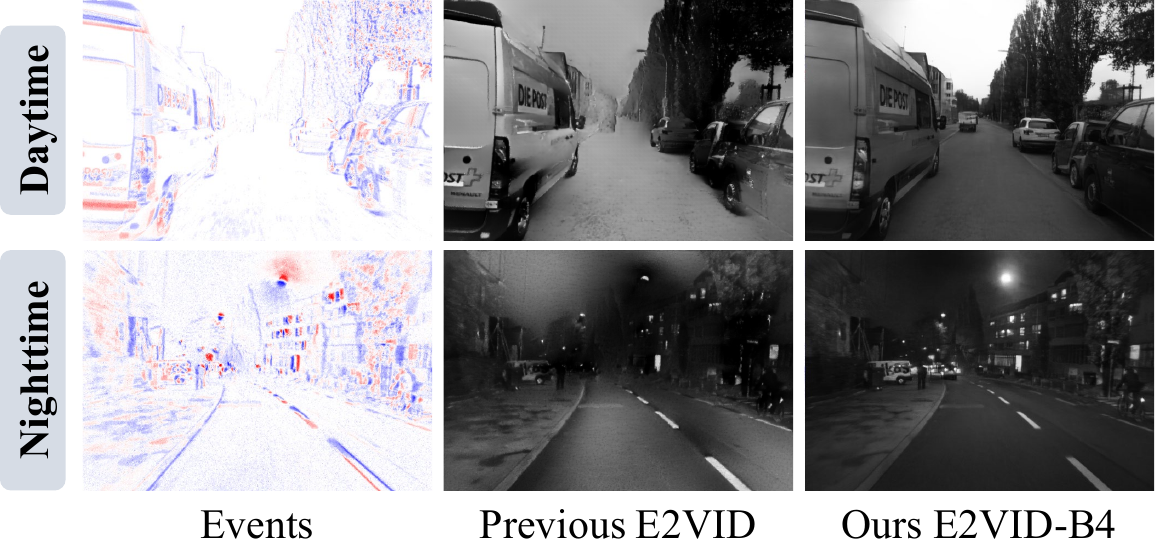}
    \end{minipage}\hfill 
    \begin{minipage}{0.42\textwidth}
        \centering 
        \includegraphics[width=\linewidth]{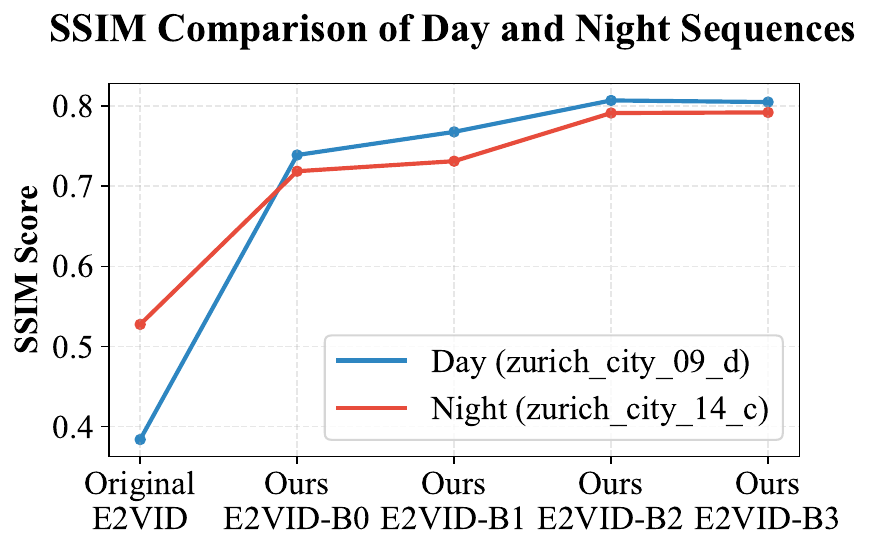}
    \end{minipage}
    % \vspace{-8pt}
    \caption{
        Left: Qualitative results of grayscale reconstruction for daytime and nighttime scenes by previous E2VID~\cite{E2VID} and our retrained E2VID-B4. 
        Right: SSIM comparison between day and night sequences for different E2VID variants.
    }
    \label{fig:e2vid} 
    % \vspace{-5pt}
\end{figure*}

\subsection{Dual Spatiotemporal Attention (DSA)}

While LTA effectively models global temporal dependencies, it may overlook subtle but critical local motions (e.g., edge displacement, texture deformation) between adjacent frames, which are vital for accurate temporal reasoning in high-speed scenarios. To bridge this gap, we propose DSA, a dual-path attention mechanism that synergizes inter-frame feature alignment and local spatiotemporal consistency modeling. This design addresses two key challenges: (1) Explicit temporal correspondence between consecutive frames to mitigate error accumulation in dynamic scenes, and (2) Preservation of motion continuity within localized spatiotemporal neighborhoods to capture transient patterns. By hierarchically integrating global and local temporal cues, DSA complements LTA to form a unified spatiotemporal representation that is both contextually aware and motion-sensitive.

\noindent \textbf{Inter-Frame Cross-Attention.} 
This component models temporal dependencies between consecutive frames through cross-attention operations.
Query vectors are computed from previous frame $f_{t-1}$ whereas key and value vectors are derived from current frame $f_t$:
\begin{equation}
Q = W^{Q} \cdot f_{t-1},~
K = W^{K} \cdot f_{t},~
V = W^{V} \cdot f_{t}.
\end{equation}
Then, the cross-attention mechanism generates adaptive feature updates through the similar operation in equation~\ref{equ:attenion}.
Our inter-frame cross-attention enables dynamic feature alignment between adjacent frames, thereby capturing motion patterns and appearance variations through learnable attention weights.

\noindent \textbf{Local Window Self-Attention.}
To complement global temporal modeling, this operator is designed to focus on spatiotemporal consistency within local neighborhoods. For each spatial location $(h,w)$ in frame $t$, we construct a spatiotemporal window $\Omega_{t-1}^{(h,w)}$ that spans:
\begin{equation}
\Omega^{h,w}_{t-1} = \{f^{i,j}_{t-1} \mid i \in [h-\delta, h+\delta], j \in [w-\delta, w+\delta] \},
\end{equation}
where $\delta$ defines the window size. Then, query, key and value vectors in self-attention operation are obtained from: 
\begin{equation}
\label{equ:window_qkv}
\begin{aligned}
Q &= W^{Q} \cdot [f^{h,w}_t,\Omega^{h,w}_{t-1}],~
K = W^{K} \cdot [f^{h,w}_t,\Omega^{h,w}_{t-1}], \\
V &= W^{V} \cdot [f^{h,w}_t,\Omega^{h,w}_{t-1}].
\end{aligned}
\end{equation}
This hierarchical design enables simultaneous modeling of fine-grained local motions and global temporal dependencies through complementary attention pathways.

\subsection{Deep Feature Guidance Mechanism (DFGM)}

While LTA and DSA enable spatiotemporal feature aggregation, their reliance on shallow temporally propagated features risks semantic drift, as they lack the semantic stability required for robust temporal reasoning. To address this, DFGM introduces semantic persistence by adaptively fusing high-level, task-specific features (e.g., object boundaries in segmentation, depth discontinuities in estimation) from historical frames into the temporal fusion process. These deep features, extracted from the VFM’s decoder layers, provide anchor points of semantic consistency across time steps, guiding the model to prioritize temporally invariant attributes (e.g., object identity, material properties) while aggregating motion-sensitive shallow features. By bridging the semantic hierarchy, DFGM ensures that temporal fusion aligns not only geometrically but also semantically, mitigating error propagation in dynamic scenarios.

Specifically, our DFGM bridges semantic hierarchies by fusing high-level guidance signals with temporal feature streams. Let $\mathbf{F}_{t-1:t-k}$ denote semantic-rich features from frame $t-1$ to $t-k$ in the VFM's decoder layers. We first project these features into the temporal fusion token space via patch-wise embedding:
\begin{equation}
\mathbf{G}_{t-1:t-k} = \mathrm{PatchEmbed}(\mathbf{F}_{t-1:t-k}) \in \mathbb{R}^{H\times W\times C},
\end{equation}
where $\mathrm{PatchEmbed}(\cdot)$ is implemented by strided non-overlapping convolution. These guidance features are then additively fused with the historical shallow features $f_{t-1:t-k}$ across the temporal dimension:
\begin{equation}
\tilde{f}_{t-1:t-k} = f_{t-1:t-k} + \mathbf{G}_{t-1:t-k}.
\end{equation}
The enriched features $\tilde{f}_{t-1:t-k}$ then replace original shallow features $f_{t-1:t-k}$ in both long-range temporal attention (equation~\ref{equ:temporal_qkv}) and local window self-attention (equation~\ref{equ:window_qkv}) computations.
Our DFGM injects semantic persistence by propagating class-discriminative patterns across time steps.

\subsection{Other Improvements}
\label{otherimprovement}
\noindent \textbf{High-Fidelity Reconstruction by retrained E2VID.} 
Existing E2VID models~\cite{E2VID,E2VID2,SPADE} employ the spatial encoder-decoder in a U-Net-style~\cite{UNet} architecture with recurrent components. They are trained on low-resolution synthetic datasets from event simulators~\cite{ESIM} and exhibit significant performance degradation in real-world scenarios. Thus, we retrain multiple E2VID variants (B0-B5) specifically optimized for the DSEC benchmark with progressive increases in encoder depth and channel dimensions, detailed in Section~\ref{sec:e2vid_arch}. Quantitative comparisons in Figure~\ref{fig:e2vid} demonstrate significant SSIM improvements over original E2VID, with visual results showing enhanced edge preservation and noise suppression in Figure~\ref{fig:e2vid}. Notably, our training uses only daytime DSEC sequences yet achieves robust night scene reconstruction.

\noindent \textbf{Zero-Initialized Residual Connections.}
To preserve the pretrained VFM knowledge while integrating temporal context fusion, we adopt a zero-initialized residual connection strategy inspired by ControlNet~\cite{ControlNet}. For a standard ViT block with output feature $f_\text{out}=\text{ViT}(f_\text{in})$, our temporal context fusion block $\mathcal{T}$ is integrated as:
\begin{equation}
\tilde{f}_\text{out} = f_\text{out} + \text{Linear}(\mathcal{T}(f_\text{in})),
\end{equation}
where the weight and bias in the linear layer ($\text{Linear}$) are initialized to zero, which ensures $\mathcal{T}$ initially behaves as an identity function, minimally perturbing the original VFM behavior during early training stages.

\noindent \textbf{Parameter Sharing.}
To balance performance and computational efficiency, we share parameters across temporal context fusion blocks at different layers. Specifically, the attention matrices, feed-forward networks, and patch embedding in DFGM across different temporal blocks share the same parameters. 
This reduces parameters by about 75\% compared to independent blocks, while maintaining competitive performance, as demonstrated in Section~\ref{sec:ablation}.

\begin{figure*}[t]
    \centering
    \includegraphics[width=0.99\textwidth]{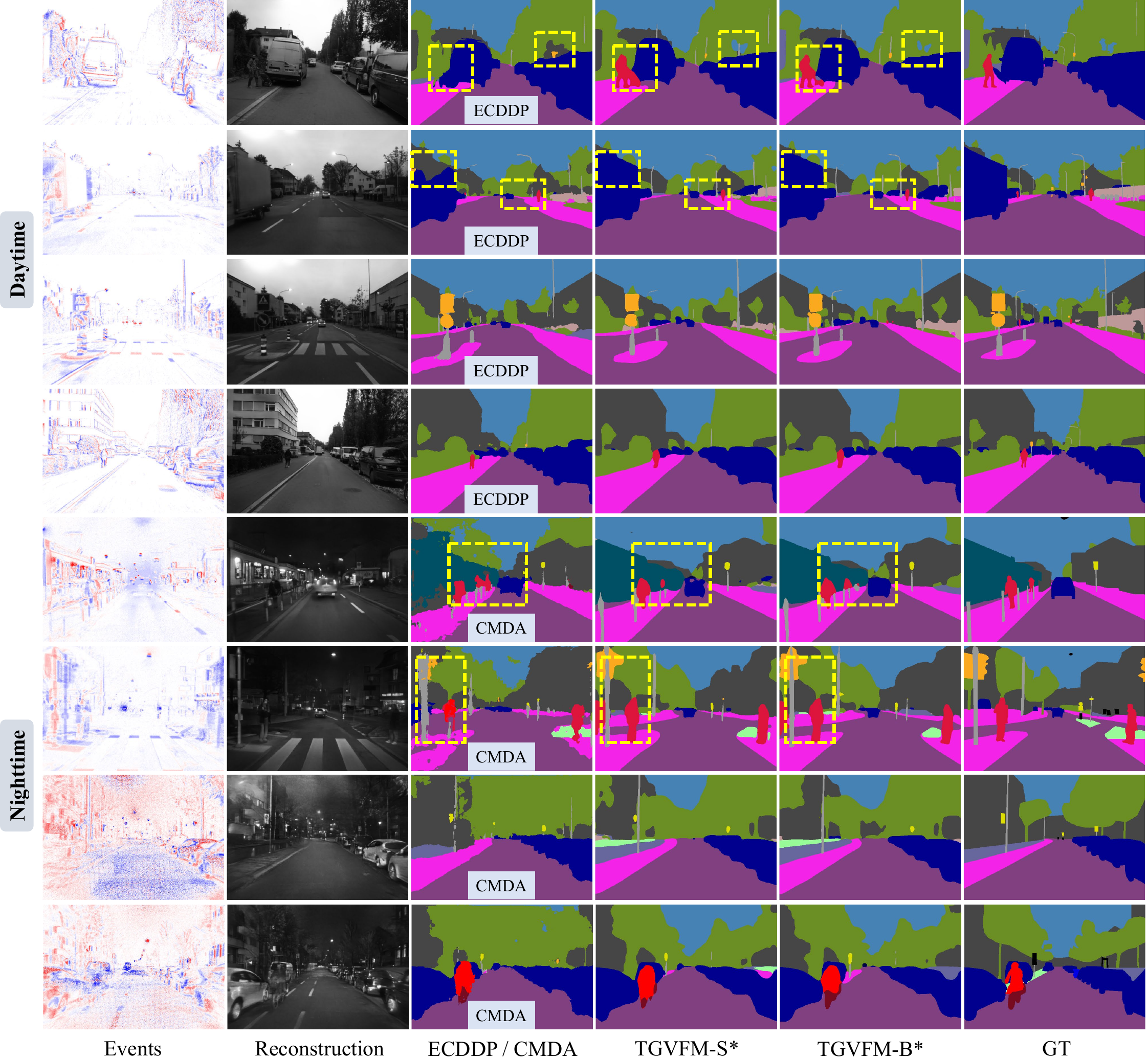}
    % \vspace{-8pt}
    \caption{
        Comparison results of semantic segmentation with ECDDP~\cite{ECDDP} (Daytime) and CMDA~\cite{CMDA} (Nighttime).
    }
    \label{fig:seg_vis}
    % \vspace{-5pt}
\end{figure*}

% Per-class Semantic Segmentation
\begin{table*}[t]
\caption{Quantitative semantic segmentation results evaluated with MIoU (\%) on the DSEC-Seg-Day~\cite{ESS} Dataset. $\dag$ We retrain ECDDP~\cite{ECDDP} to obtain the IoU metrics for each category and the original MIoU in ECDDP is 61.25. The \firstone{best} and \secondone{second best} are highlighted.}
    \centering
    \resizebox{0.99\linewidth}{!}{
    \setlength{\tabcolsep}{1.5mm}
    \begin{tabular}{c|c|ccccccccccc|c}
    \toprule
    \textbf{Method} & \textbf{\shortstack{Param.\\(M)}} & \rotatebox{90}{Sky} & \rotatebox{90}{Build.} & \rotatebox{90}{Fence} & \rotatebox{90}{Person} & \rotatebox{90}{Pole} & \rotatebox{90}{Road} & \rotatebox{90}{S.walk} & \rotatebox{90}{Veg.} & \rotatebox{90}{Car} & \rotatebox{90}{Wall} & \rotatebox{90}{Tr. S.} & \multicolumn{1}{c}{\textbf{\shortstack{Avg. \\ MIoU$\uparrow$}}} \\
    \midrule
    \midrule
    ESS~\cite{ESS} & 17.4 & 93.73 & 76.72 & 14.79 & 28.55 & 27.38 & 92.24 & 63.74 & 78.78 & 69.55 & 11.42 & 30.82 & 53.43 \\
    HMNet~\cite{HMNet} & 25.7 & 93.48 & 79.49 & 18.02 & 28.12 & 29.99 & 93.50 & 68.62 & 81.87 & 79.79 & 14.08 & 40.98 & 57.09 \\
    OpenESS~\cite{OpenESS} & - & - & - & - & - & - & - & - & - & - & - & - & 57.21 \\
    ECDP~\cite{ECDP} & 54.4 & - & - & - & - & - & - & - & - & - & - & - & 59.16 \\
    ECDDP~$\dag$~\cite{ECDDP} & 59.9 & 95.41 & 81.47 & 16.11 & 31.99 & 35.69 & 93.94 & 71.08 & 84.51 & 80.49 & 20.85 & 47.85 & 59.94 \\
    \midrule
    TGVFM-S & 55.3 & \secondone{96.09} & 86.19 & 26.79 & 44.00 & \secondone{41.27} & 94.89 & 74.04 & 87.12 & 87.49 & 41.05 & 48.32 & 66.11 \\
    TGVFM-S* & 55.3 & 95.46 & 86.26 & 29.23 & 49.79 & 33.76 & 95.19 & 75.18 & \secondone{88.01} & 87.20 & \secondone{47.16} & 49.62 & 66.99 \\
    TGVFM-B & 135.4 & \firstone{96.31} & \secondone{87.08} & \secondone{30.86} & \secondone{51.43} & \firstone{45.49} & \firstone{95.39} & \firstone{76.22} & 87.87 & \firstone{88.84} & 41.26 & \secondone{53.41} & \secondone{68.56} \\ 
    TGVFM-B* & 135.4 & 95.72 & \firstone{87.28} & \firstone{33.17} & \firstone{53.71} & 36.79 & \secondone{95.30} & \secondone{75.66} & \firstone{88.41} & \secondone{88.22} & \firstone{48.62} & \firstone{57.40} & \firstone{69.12} \\
    \bottomrule
    \end{tabular}
    }
    \label{tab:DSEC_Day}
\end{table*}

\begin{table*}[t]
\caption{Quantitative semantic segmentation results evaluated with MIoU (\%) on the DSEC-Seg-Night~\cite{CMDA} Dataset. As nighttime annotations are unavailable for finetuning, we can only report the distilled results. Both of TGVFM-S* and -B* are the same model in Table~\ref{tab:DSEC_Day}.}
    \centering
    \resizebox{\linewidth}{!}{
    \setlength{\tabcolsep}{1mm}
    \begin{tabular}{c|c|cccccccccccccccccc|c}
    \toprule
    \textbf{Method} & \rotatebox{90}{\textbf{Param.}} & \rotatebox{90}{Road} & \rotatebox{90}{S.walk} & \rotatebox{90}{Build.} & \rotatebox{90}{Wall} & \rotatebox{90}{Fence} & \rotatebox{90}{Pole} & \rotatebox{90}{Tr. L.} & \rotatebox{90}{Tr. S.} & \rotatebox{90}{Veg.} & \rotatebox{90}{Terr.} & \rotatebox{90}{Sky} & \rotatebox{90}{Person} & \rotatebox{90}{Rider} & \rotatebox{90}{Car} & \rotatebox{90}{Bus} & \rotatebox{90}{Train} & \rotatebox{90}{M.bike} & \rotatebox{90}{Bike} & \multicolumn{1}{c}{\textbf{\shortstack{Avg. \\ MIoU$\uparrow$}}} \\
    \midrule
    \midrule
    EV-WSSS~\cite{EV-WSSS} & 16.4 & 86.2 & 39.4 & 41.1 & 19.7 & 2.7 & 19.0 & 14.2 & - & 51.8 & - & 73.8 & 13.2 & - & 39.1 & - & - & - & - & 36.4  \\
    UDNET~\cite{UDNET} & 38.1 & - & - & - & - & - & - & - & - & - & - & - & - & - & - & - & - & - & - & 39.6  \\
    CMDA~\cite{CMDA} & 85.1 & 90.8 & 50.9 & 59.1 & 30.5 & \secondone{4.4} & 26.2 & \secondone{28.1} & 41.6 & 53.5 & 49.6 & 68.3 & 33.9 & 30.2 & 68.0 & 65.5 & 57.3 & 41.9 & 28.6 & 46.0 \\
    \midrule
    TGVFM-S* & 55.3 & \secondone{94.0} & \secondone{66.3} & \secondone{60.2} & \secondone{41.1} & 4.0 & \secondone{43.7} & 23.1 & \secondone{53.9} & \secondone{60.3} & \secondone{61.6} & \secondone{83.5} & \secondone{50.6} & \secondone{32.0} & \secondone{79.0} & \secondone{77.7} & \secondone{75.9} & \firstone{54.6} & \secondone{29.5} & \secondone{55.1} \\
    TGVFM-B* & 135.4 & \firstone{94.6} & \firstone{68.4} & \firstone{63.8} & \firstone{42.1} & \firstone{6.3} & \firstone{46.8} & \firstone{37.7} & \firstone{61.7} & \firstone{61.9} & \firstone{63.9} & \firstone{84.3} & \firstone{56.6} & \firstone{44.8} & \firstone{82.1} & \firstone{81.2} & \firstone{81.2} & \secondone{52.4} & \firstone{51.0} & \firstone{60.0} \\
    \bottomrule
    \end{tabular}
    }
    \label{tab:DSEC_Night}
\end{table*}

\section{Training}
\label{sec:loss}

Since Transform-based VFMs offer more robust generalization than early CNN-based VFMs~\cite{TransformerRobustness}, we instantiate our TGVFM with three SoTA Transformer-based VFMs: Rein~\cite{Rein} for semantic segmentation, Metric3D~\cite{Metric3D} for depth estimation, and Swin~\cite{Swin} trained on BDD100K~\cite{BDD100K} for object detection. See more results of different VFM in Section~\ref{sec:vfm_analy}.

For semantic segmentation and depth estimation, our TGVFM-S/B employs ViT-S/B~\cite{ViT} as the backbone architecture. In object detection, TGVFM-S utilizes Swin-S~\cite{Swin} as its backbone.
Our TGVFM is developed under two distinct training paradigms: the base model employs supervised learning with ground truth annotations, while TGVFM* utilizes a cross-modality distillation approach with pseudo-labels generated by a large-scale VFM taking images as input.

\subsection{Supervised Loss (TGVFM)}

\noindent \textbf{Semantic Segmentation.} We adopt the standard cross-entropy loss to optimize pixel-wise classification. For an image with  pixels and  semantic categories, the loss is computed as:
\begin{equation}
\mathcal{L}_{\text{seg}} = -\frac{1}{N} \sum_{i=1}^{N} \sum_{c=1}^{C}{y_{i,c} \log p_{i,c}},
\end{equation}
where $p_{i,c}$ denotes the predicted probability of pixel $i$ belonging to class $c$, and $y_{i,c}$ is the ground-truth one-hot label.

\noindent \textbf{Depth Estimation.} To handle scale ambiguity and logarithmic depth distribution, we employ the Scale-invariant Logarithmic (SiLog) loss~\cite{SiLog}:
% \footnote{\url{https://arxiv.org/abs/1406.2283}}:
\begin{equation}
\label{equ:silog}
\mathcal{L}_{\text{depth}} = \sqrt{\frac{1}{n} \sum_{i} g_i^2 - \frac{\lambda}{n^2} \left(\sum_{i} g_i\right)^2},
\end{equation}
where $g_i= \log d_i - \log \hat{d}_i$, $d_i$ and $\hat{d}_i$ represent predicted and ground-truth depths at pixel $i$, and $\lambda=0.5$ balance the scale-invariant terms.

\noindent \textbf{Object Detection.} Following Cascade R-CNN~\cite{CascadeRCNN},
% \footnote{\url{https://arxiv.org/abs/1712.00726}},
we optimize classification and bounding box regression through a multi-stage loss:
\begin{equation}
\label{equ:det}
\mathcal{L}_{\text{det}} = \sum_{k=1}^{K} \left( \mathcal{L}_{\text{cls}}^{(k)} + \mathcal{L}_{\text{box}}^{(k)} \right),
\end{equation}
where $K$ denotes the cascade stage number. The classification loss $\mathcal{L}_{\text{cls}}^{(k)}$ and box regression loss $\mathcal{L}_{\text{box}}^{(k)}$ at stage $k$ are:
\begin{equation}
\begin{aligned}
\mathcal{L}_{\text{cls}}^{(k)} &= -\sum_{j=1}^{M} y_j \log p_j^{(k)} + (1 - y_j) \log (1 - p_j^{(k)}),\\
\mathcal{L}_{\text{box}}^{(k)} &= \sum_{j=1}^{M} \mathbb{I}(y_j=1) \left\Vert \mathbf{b}_j^{(k)} - \mathbf{\hat{b}}_j \right\Vert_1,
\end{aligned}
\end{equation}
where $p_j^{(k)}$ and $\mathbf{b}_j^{(k)}$ are predicted class probability and bounding box coordinates for proposal $j$, $\mathbf{\hat{b}}_j$ denotes the ground-truth box, and $\mathbb{I}(\cdot)$ is an indicator function.

\subsection{Distilled Loss (TGVFM*)}
Our variant TGVFM* employs cross-modality distillation, where a larger-scale VFM processes RGB images as the teacher network to generate pseudo-labels, supervising the student network that takes reconstructed grayscale frames as input. The distillation losses are formulated as follows:

\noindent \textbf{Semantic Segmentation.}
We apply L1 loss between the final probability outputs of the teacher and student networks, which preserves channel-wise information more effectively than cross-entropy.
% \begin{equation}
% \mathcal{L}_{\text{seg}}^{\text{distill}} = \frac{1}{N \cdot C} \sum_{i=1}^{N} \sum_{c=1}^{C} |p_{i,c}^T - c|,
% \end{equation}
% where $p_{i,c}^T$ and $p_{i,c}^S$ denote the probability outputs of the teacher and student networks.

\noindent \textbf{Depth Estimation.}
The student network is optimized using the SiLog loss in equation~\ref{equ:silog} between its predictions and the teacher-generated pseudo-labels.

\noindent \textbf{Object Detection.}
Pseudo-labels are generated by filtering teacher-produced bounding boxes with confidence scores above 0.4. The student network is trained using the same detection loss in equation~\ref{equ:det}, computed against pseudo-labels.

\section{Experiments}
\label{sec:exper}

\noindent \textbf{Datasets.} 
Our approach is evaluated on the DSEC dataset~\cite{DSEC}, a comprehensive urban driving benchmark that significantly advances multi-modality perception research. DSEC uniquely integrates three complementary sensing modalities: high-resolution event-based data, synchronized high-quality RGB images, and dense LiDAR points.
It includes multiple tasks such as semantic segmentation~\cite{ESS}, monocular depth estimation~\cite{DSEC}, and object detection~\cite{DSEC-Det}.

\noindent \textbf{Implementation Detail.} 
For the E2VID retraining, we employ a batch size of 2, optimized over 50,000 iterations. 
In optimizing our TGVFM, we utilize a batch size of 2 across 40,000 iterations, incorporating Long-Range Temporal Attention (LTA) with a window size \(k=3\) for temporal feature aggregation. Four temporal blocks are evenly distributed across 12 ViT blocks, enabling progressive spatiotemporal integration across network depth.
Unless otherwise specified, we default to the use of the reconstruction results from E2VID-B3 for training and testing.
All experiments are conducted on a single NVIDIA L40s GPU.

\begin{figure*}[t]
    \centering
    \includegraphics[width=0.99\textwidth]{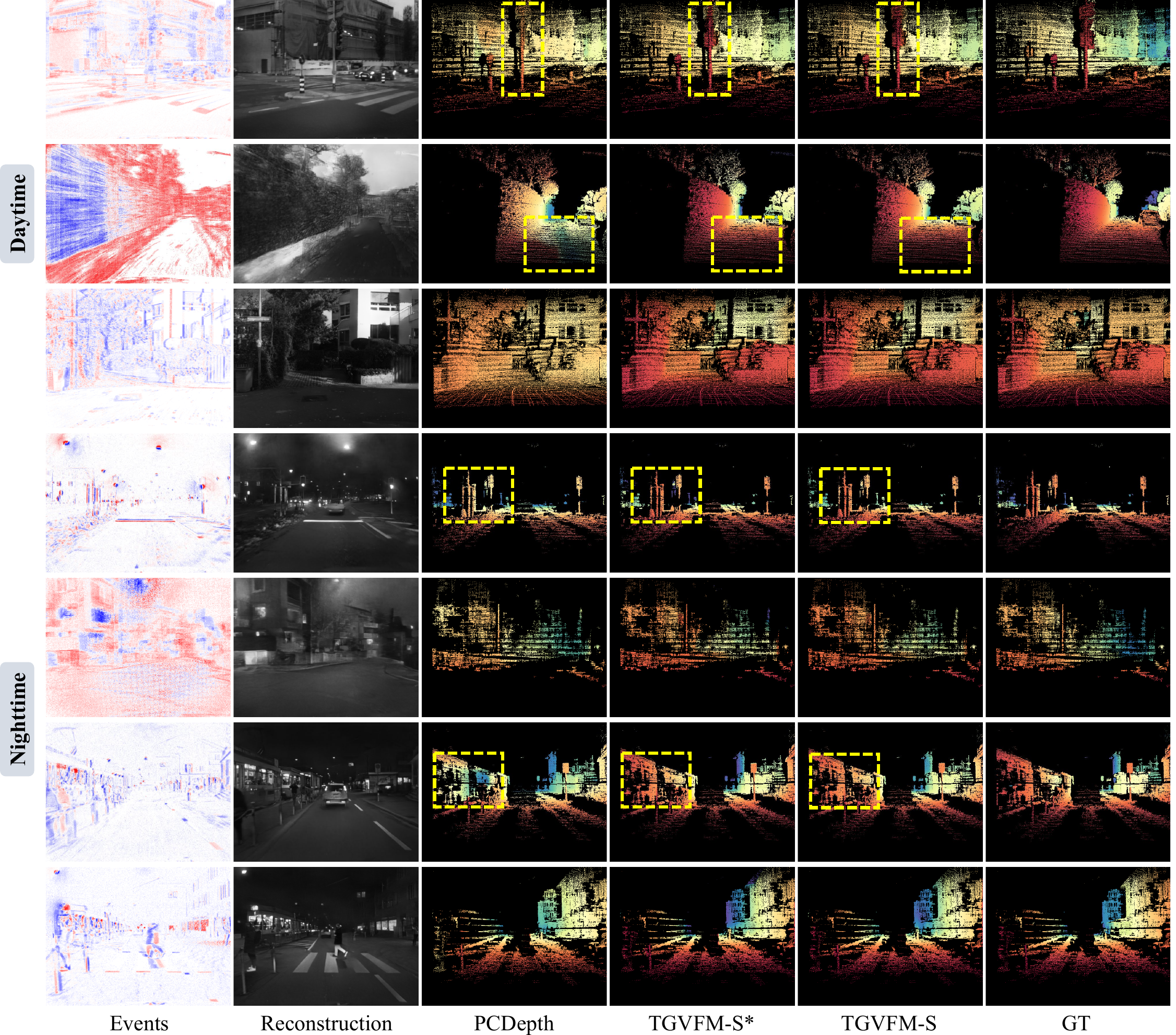}
    % \vspace{-8pt}
    \caption{
        Quantitative comparison results of monocular depth estimation with SoTA PCDepth~\cite{PCDepth}.
    }
    \label{fig:depth_vis}
    % \vspace{-5pt}
\end{figure*}

% Depth Estimation
\begin{table*}[t]
\caption{Quantitative monocular depth estimation results on the DSEC~\cite{DSEC} dataset.}
\centering
\resizebox{0.95\linewidth}{!}{\setlength{\tabcolsep}{1mm}
\begin{tabular}{c|cccccc|cccccc}
    \toprule
    \multirow{2}{*}{\textbf{Method}} & \multicolumn{6}{c|}{\textbf{Daytime sequences}} & \multicolumn{6}{c}{\textbf{Nighttime sequences}} \\
    & $\delta1\uparrow$ & $\delta2\uparrow$ & $\delta3\uparrow$ & REL$\downarrow$ & RMS$\downarrow$ & RMSlog$\downarrow$ & $\delta1\uparrow$ & $\delta2\uparrow$ & $\delta3\uparrow$ & REL$\downarrow$ & RMS$\downarrow$ & RMSlog$\downarrow$ \\
    \midrule
    \midrule
    % IDisc & 0.873 &  0.970  &  0.993   & 0.106   &   3.973   & 0.155    &0.872 & 0.970&0.990 & 0.115    &4.266  &    0.165  \\
    EReFormer~\cite{EReFormer} & 0.746 & 0.932 & 0.980 & 0.183 & 4.741 & 0.216 & 0.763 & 0.943 & 0.986 & 0.180 & 5.260 & 0.212 \\
    PCDepth~\cite{PCDepth} & 0.878 & 0.971 & \secondone{0.992} & 0.111 & \secondone{3.756} & \secondone{0.146} & 0.821 & 0.960 & 0.989 & 0.145 & 4.650 & 0.180 \\
    \midrule
    TGVFM-S* & \secondone{0.888} & \secondone{0.972} & \secondone{0.992} & \secondone{0.100} & 4.188 & 0.157 & \secondone{0.872} & \secondone{0.970} & \secondone{0.990} & \secondone{0.116} & \secondone{4.560} & \secondone{0.170} \\ % 19-1
    TGVFM-S & \firstone{0.900} & \firstone{0.976} & \firstone{0.994} & \firstone{0.092} & \firstone{3.655} & \firstone{0.140} & \firstone{0.886} & \firstone{0.974} & \firstone{0.992} & \firstone{0.109} & \firstone{4.167} & \firstone{0.157} \\ % 18-1
    \bottomrule
\end{tabular}
}
% \vspace{-5pt}
\label{tab:DSEC-Depth}
\end{table*}

\subsection{Comparison with SoTAs.}

\noindent \textbf{Semantic Segmentation.} 
As shown in Table~\ref{tab:DSEC_Day} and~\ref{tab:DSEC_Night}, our method achieves SoTA performance on both DSEC-Seg-Day~\cite{ESS} and DSEC-Seg-Night~\cite{CMDA} datasets. For daytime segmentation, our distilled TGVFM-S/B*, achieves a remarkable 66.99\%/69.12\% MIoU, outperforming previous best methods ECDDP (59.94\%) by significant margins. Notably, we observe consistent improvements across 11 object categories, particularly for challenging classes such as Fence (+17.1\%), Person (+21.7\%), and Wall (+27.8\%), demonstrating superior feature discrimination.

The nighttime evaluation in Table~\ref{tab:DSEC_Night} reveals even more pronounced advantages, where TGVFM-B* achieves 60.0\% MIoU without any nighttime-specific labels for fine-tuning, surpassing CMDA by 14\%. 
Qualitative results in Figure~\ref{fig:seg_vis} demonstrate enhanced segmentation precision, especially for moving vehicles and fine-grained structures in the daytime. Further, in the nighttime, our TGVFM* demonstrates superior robustness and achieves a significant reduction in artifacts.

\noindent \textbf{Monocular Depth Estimation.} 
Our framework sets new benchmarks for event-based monocular depth estimation as evidenced in Table~\ref{tab:DSEC-Depth}. TGVFM-S achieves unprecedented performance with 0.092 REL on daytime sequences, reducing PCDepth's errors by 17.1\%. More crucially, our distilled model TGVFM-S* maintains strong nighttime robustness (0.116 REL), outperforming PCDepth by 20\%. The consistent $\delta1$-$\delta3$ metrics across both sequences confirm our architecture's inherent capability to handle illumination variations through spatiotemporal feature fusion.

Visual comparisons in Figure~\ref{fig:depth_vis} offer a compelling demonstration of the superiority of our approach. In the upper panel, our method exhibits enhanced robustness to rapid camera motion, accurately recovering fine-grained details such as traffic light edges with remarkable sharpness. The lower part highlights our TGVFM's capability to reliably estimate depth in challenging low-light conditions, as evidenced by precise depth predictions for a train.

\noindent \textbf{Object Detection.} 
As presented in Table~\ref{tab:dsec-det-table}, our TGVFM achieves 47.7\% mAP on DSEC-Det, surpassing previous event-based detectors by 6.6\% to 9.0\%. The substantial improvements in AP$_{\mathbf{50}}$ (74.1\%) and AP$_{\mathbf{75}}$ (51.3\%) indicate precise localization capabilities, particularly for medium-sized objects where we attain +16.9\% higher than SSM.  

The unified performance across three tasks underscores our TGVFM's versatility in processing event-based data. The consistent daytime-nighttime superiority further highlights its robust under dynamic lighting conditions.

\begin{table}[t]
\caption{Comparative study of SoTA event camera detectors on the DSEC-Det~\cite{DSEC-Det} dataset.}
\centering
\resizebox{0.99\linewidth}{!}{\setlength{\tabcolsep}{1mm}
\begin{tabular}{c|cccccc}
    \toprule
    \textbf{Method} & \textbf{mAP$\uparrow$} & \textbf{AP}$_{\mathbf{50}}$$\uparrow$ & \textbf{AP}$_{\mathbf{75}}$$\uparrow$ & \textbf{AP}$_{\mathbf{S}}$$\uparrow$ & \textbf{AP}$_{\mathbf{M}}$$\uparrow$ & \textbf{AP}$_{\mathbf{L}}$$\uparrow$
    \\
    \midrule
    \midrule
    RVT~\cite{RVT} & 38.4 & 58.7 & 41.3 & 29.5 & 50.3 & \firstone{81.7} \\
    SAST~\cite{SAST} & 38.1 & 60.1 & 40.0 & 29.8 & 48.9 & \secondone{79.7} \\
    SSM~\cite{SSM} & 38.0 & 55.2 & 40.6 & 28.8 & 52.2 & 77.8 \\
    LEOD~\cite{LEOD} & \secondone{41.1} & 65.2 & \secondone{43.6} & \secondone{35.1} & 47.3 &73.3 \\
    \midrule
    TGVFM-S* & 40.0 & \secondone{72.7} & 38.6 & 27.4 & \secondone{62.7} & 74.3 \\ % 22
    TGVFM-S & \firstone{47.7} & \firstone{74.1} & \firstone{51.3} & \firstone{35.6} & \firstone{69.1} & 78.6 \\ % 22
    \bottomrule
\end{tabular}
}
% \vspace{-5pt}
\label{tab:dsec-det-table}
\end{table}

\begin{table}[t]
\caption{Ablation study on the DSEC-Seg-Day~\cite{ESS} dataset.}
\centering
\resizebox{0.7\linewidth}{!}{\setlength{\tabcolsep}{1mm}
\begin{tabular}{ccc|cc}
    \toprule
    \textbf{LTA} & \textbf{DSA} & \textbf{DFGM} & \textbf{MIoU$\uparrow$} & \textbf{Impro.$\uparrow$} \\
    \midrule
    \midrule
               &            &            & 65.88 & -      \\
    \Checkmark &            &            & 67.89 & + 2.01 \\
    \Checkmark &            & \Checkmark & 68.33 & + 2.45 \\
               & \Checkmark &            & 67.81 & + 1.93 \\
               & \Checkmark & \Checkmark & 68.11 & + 2.23 \\
    \Checkmark & \Checkmark &            & 68.13 & + 2.25 \\   
    \Checkmark & \Checkmark & \Checkmark & 69.01 & + 3.13 \\
    \bottomrule
\end{tabular}
}
% \vspace{-5pt}
\label{tab:ablation}
\end{table}

\begin{figure}[t]
    \centering
    \includegraphics[width=0.48\textwidth]{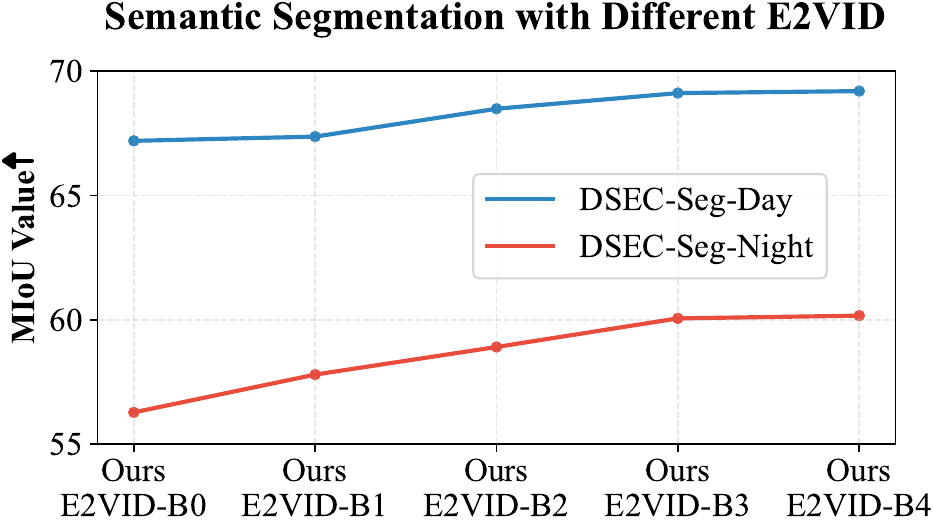}
    % \vspace{-8pt}
    \caption{
        Evaluation of integrating TGVFM-B* with E2VID on DSEC-Seg-Day~\cite{ESS} and -Night~\cite{CMDA}.
    }
    \label{fig:e2vid_seg}
    % \vspace{-5pt}
\end{figure}

\begin{figure}[t]
    \centering
    \includegraphics[width=0.48\textwidth]{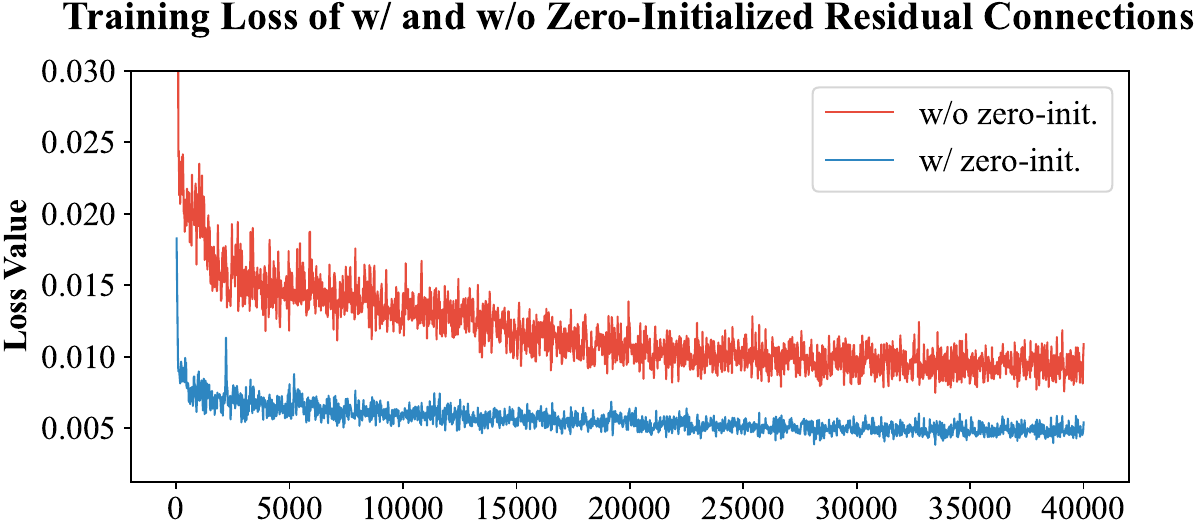}
    % \vspace{-8pt}
    \caption{
        Impact of zero-initialized residual Connections on training losses.
    }
    \label{fig:zero_init_loss}
    % \vspace{-5pt}
\end{figure}

\subsection{Ablation Study}
\label{sec:ablation}
Our comprehensive ablation in Table~\ref{tab:ablation} assesses contributions of each component in the TGVFM-B architecture with E2VID-B4.
When applying only the LTA, we observe a 2.01\% MIoU improvement, demonstrating its effectiveness in capturing extended temporal dependencies. The complementary benefits of DFGM become evident when combined with LTA, resulting in additional 0.44\% gains.
Notably, the DSA module independently achieves 1.93\% improvement, highlighting its capacity for joint spatiotemporal modeling. The synergistic combination of DSA with DFGM further enhances performance (+2.23\%).
Crucially, the unified framework with all components achieves the highest MIoU of 69.01\% (+3.13\%), highlighting their complementary roles in addressing distinct challenges.

\begin{table*}[t]
\caption{Effect of Memory-Bank Window Size $k$ on Segmentation Performance.}
    \centering
    \resizebox{0.99\linewidth}{!}{
    \setlength{\tabcolsep}{1.5mm}
    \begin{tabular}{c|c|ccccccccccc|c}
    \toprule
    \textbf{Method} & \textbf{\shortstack{Infer.\\Time\\(ms)}} & \rotatebox{90}{Sky} & \rotatebox{90}{Build.} & \rotatebox{90}{Fence} & \rotatebox{90}{Person} & \rotatebox{90}{Pole} & \rotatebox{90}{Road} & \rotatebox{90}{S.walk} & \rotatebox{90}{Veg.} & \rotatebox{90}{Car} & \rotatebox{90}{Wall} & \rotatebox{90}{Tr. S.} & \multicolumn{1}{c}{\textbf{\shortstack{Avg. \\ MIoU$\uparrow$}}} \\
    \midrule
    \midrule
    $k{=}1$ & 28.0 & 95.19 & 85.75 & 23.89 & 49.26 & 33.03 & 94.98 & 74.01 & 87.90 & 87.08 & 47.21 & 48.15 & 66.04 \\
    $k{=}2$ & 28.2 & 95.26 & 86.15 & 27.74 & 50.12 & 33.74 & 95.14 & 75.00 & 88.08 & 87.26 & 46.27 & 48.97 & 66.70 \\
    $k{=}3$ & 28.5 & 95.46 & 86.26 & 29.23 & 49.79 & 33.76 & 95.19 & 75.18 & 88.01 & 87.20 & 47.16 & 49.62 & \firstone{66.99} \\
    $k{=}4$ & 29.2 & 95.30 & 86.35 & 27.85 & 51.01 & 33.46 & 95.10 & 74.67 & 88.01 & 87.45 & 46.34 & 49.72 & 66.84 \\
    $k{=}5$ & 29.6 & 95.35 & 86.20 & 28.04 & 50.81 & 33.19 & 95.17 & 74.92 & 88.06 & 87.46 & 45.71 & 50.95 & 66.90 \\
    \bottomrule
    \end{tabular}
    }
    \label{tab:memory}
\end{table*}

\begin{table*}[t]
\caption{Ablation on Event Representation.}
    \centering
    \resizebox{0.99\linewidth}{!}{
    \setlength{\tabcolsep}{1.5mm}
    \begin{tabular}{c|ccccccccccc|c}
    \toprule
    \textbf{Method} & \rotatebox{90}{Sky} & \rotatebox{90}{Build.} & \rotatebox{90}{Fence} & \rotatebox{90}{Person} & \rotatebox{90}{Pole} & \rotatebox{90}{Road} & \rotatebox{90}{S.walk} & \rotatebox{90}{Veg.} & \rotatebox{90}{Car} & \rotatebox{90}{Wall} & \rotatebox{90}{Tr. S.} & \multicolumn{1}{c}{\textbf{\shortstack{Avg. \\ MIoU$\uparrow$}}} \\
    \midrule
    \midrule
    TGVFM (Time Surfaces) & 94.53 & 83.40 & 25.86 & 35.65 & 28.44 & 94.40 & 72.74 & 85.67 & 83.08 & 31.76 & 36.15 & 61.06 \\
    TGVFM (Voxel Grids)   & 94.47 & 82.59 & 23.84 & 35.72 & 25.86 & 94.90 & 74.03 & 85.28 & 83.59 & 38.94 & 40.10 & 61.76 \\
    TGVFM (E2VID)         & 95.46 & 86.26 & 29.23 & 49.79 & 33.76 & 95.19 & 75.18 & 88.01 & 87.20 & 47.16 & 49.62 & \firstone{66.99} \\
    \bottomrule
    \end{tabular}
    }
    \label{tab:event_representation}
\end{table*}

\begin{figure}[h]
    \centering
    \includegraphics[width=0.48\textwidth]{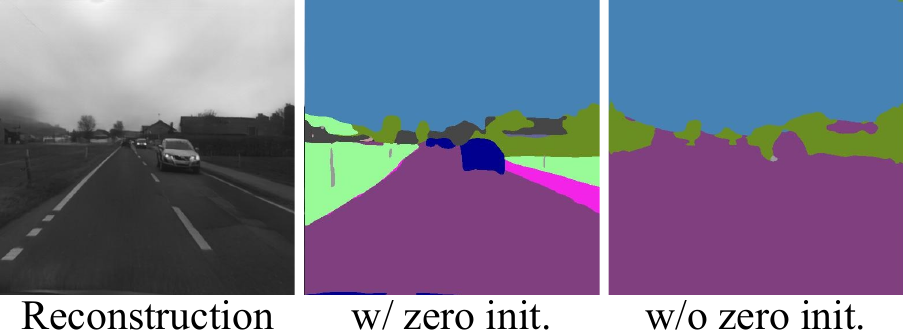}
    % \vspace{-8pt}
    \caption{
        Visualization of the 200th training iteration w/ and w/o zero-initialized residual Connections.
    }
    \label{fig:zero_init_200th}
    % \vspace{-5pt}
\end{figure}

\noindent \textbf{Zero-Initialized Residual Connections.}
Our analysis reveals the critical role of zero-initialized residual connections in stabilizing training and preserving pre-trained knowledge in VFMs. As shown in Figure~\ref{fig:zero_init_loss}, models without zero-initialized connections (w/o zero-init.) exhibit unstable optimization dynamics, characterized by high initial loss values and significant training fluctuations. In contrast, models with zero-initialized connections (w/ zero-init.) start with a lower initial loss and converge smoothly to a stable value.
The qualitative results in Figure~\ref{fig:zero_init_200th} further validate this behavior. At the 200th training iteration, the zero-initialized variant maintains the visual fidelity and reasoning capabilities of the original VFM, producing coherent predictions that align with the reconstruction semantics. Conversely, the non-zero-initialized model suffers from catastrophic forgetting of pre-trained features, generating inconsistent outputs.

\begin{table}[t]
\caption{Ablation of parameters sharing of our proposed temporal context fusion block on the DSEC-Seg-Day~\cite{ESS} dataset with E2VID-B4.}
\centering
\resizebox{0.8\linewidth}{!}{ % \setlength{\tabcolsep}{1mm}
\begin{tabular}{c|cccccc}
\toprule
\multirow{2}{*}{\textbf{Method}} & \textbf{Params.} & \textbf{TCFB} & \textbf{Avg} \\
 & \textbf{Sharing} & \textbf{Params. (M)} & \textbf{MIoU$\uparrow$} \\
\midrule
\midrule
% \Checkmark
\multirow{2}{*}{TGVFM-S*} & \XSolidBrush & 22.8 & 66.85 \\
 & \Checkmark & 5.4 & 66.73 \\
\midrule
\multirow{2}{*}{TGVFM-B*} & \XSolidBrush & 71.7 & 69.07 \\
 & \Checkmark & 19.0 & 69.20 \\
\bottomrule
\end{tabular}
}
\label{tab:param_share}
\end{table}

\noindent \textbf{Parameters Sharing.}
Given the consistent channel dimension in the ViT backbone~\cite{ViT} used for semantic segmentation (Rein~\cite{Rein}) and depth estimation (Metric3D~\cite{Metric3D}), we implement parameter sharing across across different temporal context fusion block.
This approach significantly reduces parameters in our proposed TCFB while maintaining competitive performance. 
As shown in Table~\ref{tab:param_share}, the results demonstrate two advantages of our parameter-sharing strategy: 1) A 76.3\% parameter reduction in TCFB (from 22.8M to 5.4M) for TGVFM-S* and 73.5\% reduction (71.7M to 19.0M) for TGVFM-B*, 2) Maintained segmentation performance with $<$0.2\% MIoU difference across both model scales. This validates that our parameters sharing effectively preserves temporal reasoning capability while eliminating redundant parameters.

\noindent \textbf{Memory Bank Size.}
In our framework, the memory bank plays a critical role in the Long-Range Temporal Attention (LTA) and Dual Spatiotemporal Attention (DSA) modules. However, it is important to note that DSA only utilizes the immediate past feature $f_{t-1}$. Therefore, the memory bank window size $k$ exclusively affects the temporal feature aggregation in the LTA module. To assess the sensitivity of our model to the memory-bank size, we conducted a series of experiments by varying $k$, which defines the number of past frames stored in the memory bank. This analysis highlights the trade-off between segmentation accuracy and computational complexity. Our results in Table~\ref{tab:memory} show that increasing $k$ from 1 to 3 leads to consistent performance improvements, with the best result achieved at $k=3$, where the model reaches an average mIoU of 66.99. Beyond $k=3$, we observe diminishing returns, indicating that performance gains plateau while computational overhead continues to grow. These findings suggest that a memory window size of 3 provides an effective balance between efficiency and long-range temporal modeling capability. This sensitivity analysis reinforces the robustness of our component design and the practical effectiveness of our memory-based temporal attention mechanism.

\noindent \textbf{Event Representation.}
To investigate the influence of event data representation, we compare our default E2VID-based reconstruction with alternative formats such as voxel grids and time surfaces. E2VID converts asynchronous events into image-like grayscale frames, allowing direct compatibility with pretrained VFMs. In contrast, voxel and time-surface representations are non-visual and require the model to learn low-level spatial semantics from scratch, thus limiting the benefit of pretrained knowledge transfer. As shown in Table~\ref{tab:event_representation}, replacing E2VID with these alternatives leads to a substantial drop in segmentation accuracy, confirming that image-domain reconstruction provides a more effective bridge between event streams and image-pretrained VFMs. This result highlights the critical role of E2VID in preserving both temporal structure and compatibility with VFMs, ultimately enabling stronger generalization across event-based tasks.

\subsection{Impact of Different E2VID}

Our analysis reveals a critical insight: increasing E2VID's model capacity yields diminishing return in TGVFM performance. As shown in Figure~\ref{fig:e2vid_seg}, MIoU on DSEC-Seg-Day saturates at 69.12\% for B3 (6.8M Params), with a marginal improvement of only 0.08\% improvement when scaling to B4 (52.0M Params). 
Similar saturation occurs in nighttime (60.06\% vs. 60.17\% for B3 and B4). This suggests that while deeper architectures enhance frame reconstruction quality, the perceptual gains become negligible.
Notably, B3 achieves comparable performance to B4 with 8× fewer parameters, demonstrating that our TGVFM effectively compensates for moderate reconstruction artifacts. The shallow B0-B2 variants (0.3M–4.2M Params) still attain competitive accuracy within 1.5\% of B4, proving our TGVFM's robustness to E2VID variations.

\begin{figure}[t]
    \centering
    \includegraphics[width=0.49\textwidth]{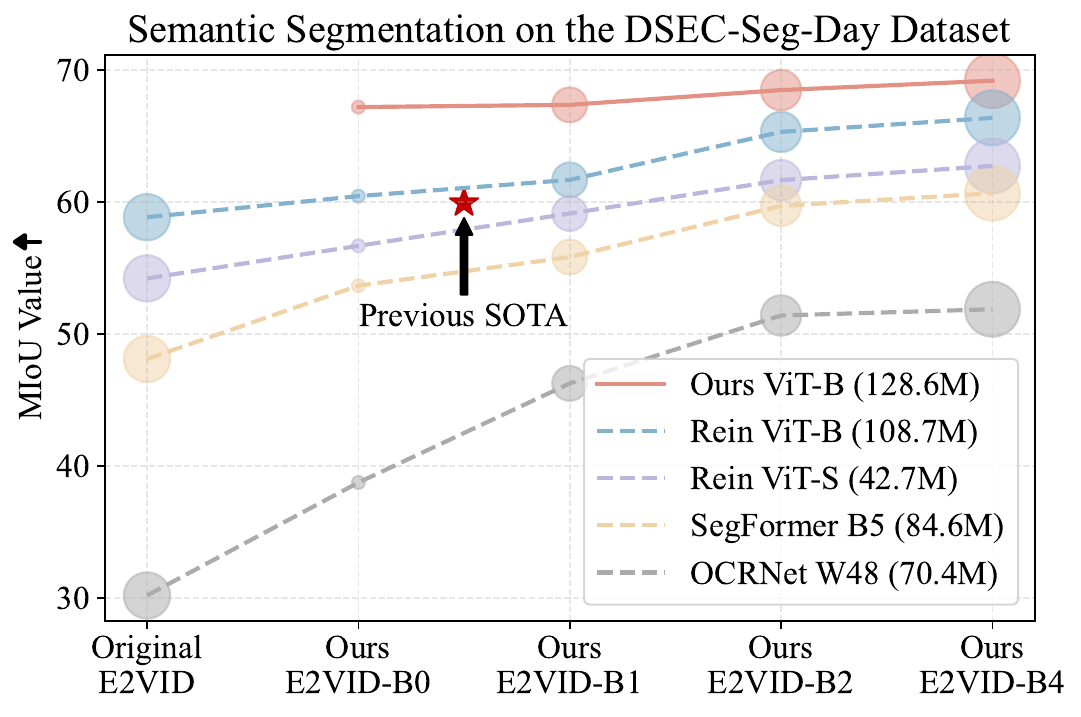}
    % \vspace{-8pt}
    \caption{
        Semantic Segmentation Performance on the DSEC-Seg-Day~\cite{ESS} Dataset with different VFM (OCRNet~\cite{OCRNet}, SegFormer~\cite{SegFormer}, and Rein~\cite{Rein}) and E2VID. The relative parameters of E2VID is expressed in terms of the size of the points.
    }
    \label{fig:vfm_analy}
    % \vspace{-5pt}
\end{figure}

\subsection{VFM Analysis}
\label{sec:vfm_analy}

In Figure~\ref{fig:vfm_analy}, we present a comprehensive evaluation with different VFM and E2VID. The horizontal axis details the E2VID reconstruction network utilized, ranging from the ``Original E2VID"~\cite{E2VID} to four of our progressively enhanced E2VID variants. 
The dashed lines depict the zero-shot capabilities of various VFMs when applied to the video outputs generated by these E2VID methods.
We benchmark three distinct classes of VFMs to understand the evolution of their representational power: (1) OCRNet W48~\cite{OCRNet} (70.4M parameters), a representative model from the CNN era; (2) SegFormer B5~\cite{SegFormer} (84.6M parameters), a widely adopted Transformer-based architecture; and (3) Rein ViT-S (42.7M parameters) and Rein ViT-B (108.7M parameters)~\cite{Rein}, which are vision transformer models leveraging powerful self-supervised DINOv2~\cite{DINOv2} backbones.
A clear trend emerges from these zero-shot evaluations: as the intrinsic capability of the VFM improves, the MIoU generally increases. This underscores the importance of the pretrained knowledge embedded within these large-scale models. 
Concurrently, the quality of the E2VID reconstruction significantly influences the final segmentation performance. Across all VFMs, a higher-quality E2VID reconstruction consistently yields better MIoU scores, demonstrating that improved event stream interpretation by E2VID is crucial for downstream tasks.

Our proposed TGVFM, represented by the solid red line, is evaluated after fine-tuning on the DSEC-Seg-Day dataset. This model integrates a Rein ViT-B backbone with our novel TCFB. As illustrated, TGVFM significantly surpasses the zero-shot performance of all considered VFMs across all E2VID variants. 
Its explicit modeling of temporal dependencies and task-specific training achieves superior segmentation accuracy.

\begin{table}[t]
\caption{Comparison of our different E2VID in terms of network architecture.}
\centering
    \resizebox{0.99\linewidth}{!}{\setlength{\tabcolsep}{1mm}
    \begin{tabular}{c|cccccc}
    \toprule
    \textbf{E2VID} & \textbf{Recurrent} & \textbf{Base} & \multirow{2}{*}{\textbf{Encoders}} & \textbf{Block} & \textbf{Residual} & \textbf{Param.} \\
    \textbf{Type} & \textbf{Block} & \textbf{Chann.} &  & \textbf{Channels} & \textbf{Blocks} & \textbf{(M)} \\
    \midrule
    \midrule
    B0 & ConvGRU & 12 & 2 & [24, 48] & 1 & 0.3 \\
    B1 & ConvGRU & 16 & 3 & [32, 64, 128] & 1 & 2.0 \\
    B2 & ConvLSTM & 20 & 3 & [40, 80, 160] & 2 & 4.2 \\
    B3 & ConvLSTM & 32 & 3 & [64, 100, 200] & 2 & 6.8 \\
    B4 & ConvLSTM & 32 & 4 & [64, 150, 300, 512] & 3 & 52.0 \\
    \bottomrule
    \end{tabular}
    }
\label{tab:e2vid}
\end{table}

\subsection{E2VID Architecture}
\label{sec:e2vid_arch}

Table~\ref{tab:e2vid} systematically compares architectural configurations of our proposed E2VID variants (B0-B4). The models progressively scale in complexity through three key dimensions: (1) recurrent block type (ConvGRU vs. ConvLSTM), (2) encoder depth (2-4 hierarchical stages with channel expansion [24$\rightarrow$512]), and (3) residual blocks (1-3 layers). B0-B3 maintain compact designs ($<$7M params) through efficient channel allocations, while B4 employs aggressive width scaling for high-performance scenarios.

\subsection{Limitations}
\label{sec:limitations}
While TGVFM achieves task-specific state-of-the-art results, its current implementation requires separate VFM backbones for different tasks, limiting its unified processing capabilities. Future research could explore to enable a single VFM backbone to handle multiple event-based tasks simultaneously.

\section{Conclusion}
\label{sec:conclusion}
This work pioneers effective integration of VFM into event-based vision. Our proposed TGVFM framework introduces a plug-and-play temporal context fusion block that enables VFMs to capture spatiotemporal dependencies without compromising pretrained knowledge. By combining long-range attention, multi-scale temporal reasoning, and deep semantic guidance, our method unlocks the potential of VFMs for event data, eliminating the need for task-specific engineering. Extensive experiments validate its superiority, achieving SOTA results across diverse tasks.

% \section*{Acknowledgments}
% This should be a simple paragraph before the References to thank those individuals and institutions who have supported your work on this article.

\bibliographystyle{IEEEtran}
\bibliography{main}

% \newpage

 


\end{document}